\documentclass{llncs}
\usepackage{graphicx}
\usepackage[caption=false]{subfig}
\usepackage{color}
\usepackage{array}

\begin{document}

\title{
Modeling the Intra-class Variability for Liver Lesion Detection using a Multi-class Patch-based CNN
}
%
\titlerunning{?}  
%
\author{Maayan Frid-Adar\thanks{Equal Contributors}\inst{1} \and Idit Diamant\inst{*1}
 Eyal Klang\inst{2} \and Michal Amitai\inst{2} \and \\
 Jacob Goldberger\inst{3} \and Hayit Greenspan\inst{1}}

\institute{*** \\ *** \\ *** \\ ***}

%
%

 \institute{Tel Aviv University, Faculty of Engineering, Department of Biomedical Engineering, Tel Aviv, Israel\\
 \and
 Sheba Medical Center, Diagnostic Imaging Department, Tel Hashomer, Israel
 \and
 Engineering Faculty, Bar-Ilan University, Israel
 }

\maketitle              

\begin{abstract}
Automatic detection of liver lesions in CT images poses a great challenge for researchers. In this work we present a deep learning approach that models explicitly the variability within the non-lesion class
,based on prior knowledge of the data,  
 to support an automated lesion detection system.
A multi-class convolutional neural network (CNN) is proposed to categorize input image patches into
sub-categories of boundary and interior patches,  the decisions of which are fused to reach a binary lesion vs non-lesion decision.
For validation of our system, we use CT images of 132 livers and 498 lesions. Our approach shows highly improved detection results that outperform the state-of-the-art fully convolutional network. Automated computerized tools, as shown in this work, have the potential in the future to support the radiologists towards improved detection.

\end{abstract}

\section{Introduction}
Liver cancer is one of the predominant cancer types, accounting for more than 600,000 deaths each year.
The number of liver tumors diagnosed throughout the world is increasing at an alarming rate. Early diagnosis and treatment is the most useful way to reduce cancer deaths.
Computed tomography (CT) images are widely used for the detection and diagnosis of liver lesions.
Manual detection is a time-consuming task which requires the radiologist to search through a 3D CT scan. Thus, there is an interest and need for automated analysis tools to assist clinicians in the detection of liver metastases in CT examinations.

Automatic liver lesion detection is a very challenging, clinically relevant task due to the substantial lesion appearance variation within and between patients (in size, shape, texture, contrast enhancement) 
 In detection, both small and large lesions are weighed similarly. This is in contrast with a segmentation task that can miss small lesions and still get a high score. 

This research problem has attracted much attention in recent years. The MICCAI 2008 Grand Challenge \cite{Deng2008} provided a good overview of possible approaches mainly for segmentation. The winner of the challenge \cite{MICCAI_2008_winner} used the AdaBoost classifier to separate liver lesions from normal liver tissue based on several local image features. In more recent works, deep learning 
\cite{Shin2016,Setio2016,Greenspan_TMI_Editorial} was applied for liver lesion detection using fully convolutional networks (FCN) \cite{FCN_paper,Christ2016}. The FCN system presented in \cite{FCN_paper} showed high detection results with TPR of 88\% and 0.74 FP per liver.

\begin{figure}[!tbp]
  \centering
  \subfloat[]
  {\includegraphics[width=0.6\textwidth]{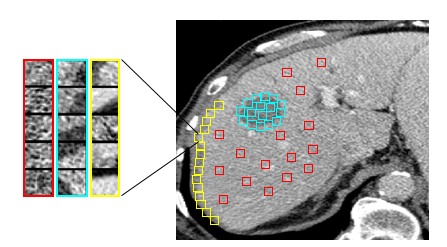}\label{Fig:patch_classes}}
  \hfill
  \subfloat[]
  {\includegraphics[width=0.4\textwidth]{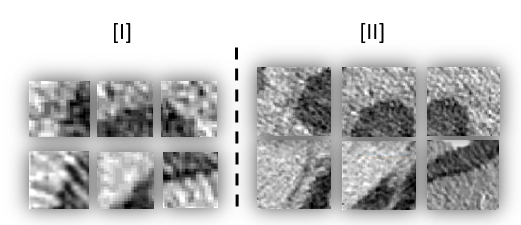}\label{Fig:Patches_small_big}}
  \caption{(a) Liver patch examples: lesion (blue), normal-interior (red) and normal- boundary (yellow);
  (b) Patch size $20\times20$ [I] and patch size $50\times50$ [II]. Top row: lesion boundary patch examples; Bottom row: liver boundary patch examples. }
\end{figure}




A deep learning approach provides a highly non-linear data representation of a given feature $x$, denoted by $h(x)$. However, at the final softmax layer, the decision is a linear  function of $h$ (it is exactly a logistic regression classifier).
 Thus, even in the most sophisticated neural network, the normal or lesion binary decision is performed by a linear classifier that is applied to $h(x)$.
 In many cases the data associated with each label is not homogeneous and is organized in clusters which have different characteristics and appearance. For example, the non-lesion patches might look different given that they either located in the liver interior or in the liver boundary.
In such cases a linear decision provided by the soft-max layer is not capable to handle the complex class structure.

	The novelty of this study is to model the solution to the task and divide the data to sub-categories correctly with specific medical task expertise. 
We demonstrate this idea by modeling the intra-class variability of the non-lesion category using a Multi-class patch-based CNN system that allocates several network classes for a single medical decision. This small modification of the standard network architecture,  provides a mean for modeling the within-class variability, yields a significant improvement over
state-of-the-art results.  Comparisons were done with state-of-the-art patch-based CNN as well as FCN.
A description of the proposed architecture is presented in Section 2. Experiments and results are shown in Section 3, followed by concluding remarks in Section 4.

%
\section{Multi-class patch-based CNN system}
%

We propose a system for lesion detection, which is based on localized patch classification into lesion vs non-lesion categories. Patches are an important representation to address the data limitation challenge, critical for modeling small and rare events such as lesions.
  Fig. \ref{Fig:patch_classes} shows examples of lesion patches, including lesion-boundary patches (shown blue zoomed-in), as well as patches from the non-lesion category, which include both normal-interior patches (red) and normal-boundary patches (yellow).  
  As can be seen in Fig. \ref{Fig:Patches_small_big}[I], liver boundary and lesion boundary areas may look alike and are difficult to distinguish when using small patches. When using a larger size window, a clearer distinction between the two categories is possible, as seen in Fig. \ref{Fig:Patches_small_big}[II]. This motivated us to
 use two different scales (patch-sizes) within the proposed system
 to capture both the local fine details and the more global spatial information.
 
 We note that the normal-interior patches and the normal-boundary patches have a distinct appearance, yet both comprise the non-lesion category. We therefore propose a solution that is aware of this intra-class variability and allocates a different label (with different soft-max parameters) for the interior and boundary patches. 
   Since eventually we are only interested in lesion/non-lesion decision, at test time we merge the non-lesion classes by summing-up their probabilities into a single probability of a non-lesion. 
   

\subsection{Patch extraction}
%

Patches are extracted from 2D liver CT scans with an expert marking for the lesions. The liver area can be segmented automatically or by an expert. The patches are labeled with their corresponding class $k$ $\in$ \{lesion, normal-interior, normal-boundary\}, automatically, according to their relative position to the boundary.
The patches are extracted around each pixel in two fields-of-view (FOV) of size $20\times20$ pixels and  $50\times50$ pixels. The patches contain localized information as well as spatial context. 
All patches are resized to $32\times32$ pixels to fit the input image size of our multi-class CNN architecture.

The amount of patches in the normal-interior class is much larger than the number of patches in other classes. Therefore, the patches are sampled randomly to balance the training set.
Data augmentation is applied to enrich the lesion class by flipping \{right,left\} and rotating in $[5,130,300]$ angles. Patches are sampled with overlap using a 2-pixel step size between patch centers.

All patches are normalized as follows: during training, the mean liver intensity was calculated on the training set: $I_{mean} = \frac{1}{N}\sum_{j}p(x_j,y_j)$ where $N$ is the number of pixels in the liver area. During the testing phase, in order to obtain uniform mean liver intensity in all livers, we shifted the patches intensities such that each test liver will have a mean intensity value equal $I_{mean}$ (mean liver intensity of training set). 
\subsection{System architecture}
\label{section:systemArc}
\begin{figure}[t]
  \centering
  {\includegraphics[width=1\textwidth]{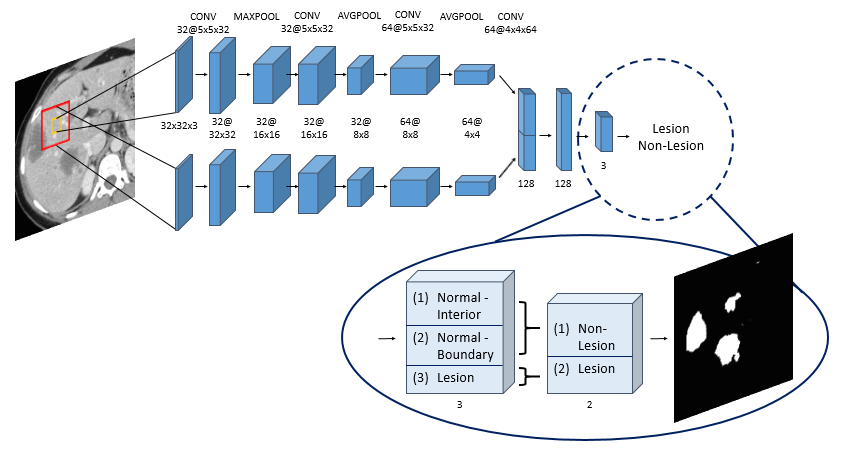}}
  \caption{The Parallel Multi-class CNN architecture: combining multi-class (to handle within class variability) and multi-scale (to handle inter-class similarity). }
  \label{Fig:CNN_System}
\end{figure}

We propose a multi-class patch-based CNN, as shown in 
Fig. \ref{Fig:CNN_System}.
The architecture of the network 
consists of 4 convolutional (conv.) layers and 3 pooling layers (one max-pooling and two avg-pooling).  Each conv. layer is followed by a ReLU activation function. The network has approx. 0.15 million parameters.

Our system receives input patches in two FOVs around each pixel.
The scaled patches are analyzed in parallel using two different networks that are late fused in the last fully-connected layer. The last layer is a softmax classifier that calculates the probabilities of the patch (which correspond to the probabilities of the center pixel) for each class. 
The probabilities are retrieved using multinomial logistic regression:
\begin{equation}
p(y=k|x)=\frac{\exp(S_k)}{\sum_{j}\exp(S_j)}
\end{equation}
where $k$ is the class label, $S_k= w_k^T h(x)+b_k$ is the score or the unnormalized log probability of class $k$ given the non-linear representation $h(x)$.

At the testing stage, the lesion detection map is generated for the interior area of the liver by replacing each pixel with its corresponding patch  probability for the lesion category.
All normal sub-category probabilities are summarized into one  non-lesion class probability:
\begin{equation}
p_{\mbox{non-lesion}} = p_{\mbox{normal-boundary}} +
p_{\mbox{normal-interior}}
\end{equation}
We threshold the lesion detection probability map to obtain a binary detection map. Herein, we term the network described above as the {\it{Parallel multi-class CNN}}. 


We also implemented another multi-class approach which is based on {\it{hierarchical binary-class}} CNNs. It is implemented by splitting the three category classification task into two steps. The first step generates a map of lesion candidates by using binary classification of lesion and non-lesion areas ('lesion detector').
The second step performs false-positive reduction using a binary-class CNN which is trained to classify between lesion and non-lesion using only normal patches located at the liver boundary and lesion patches located anywhere in the liver. This CNN is applied to the lesion candidates which were obtained from the output of the first step. We implemented the lesion detector using a binary-class CNN. 

%

%
\subsection{Training protocol}

In order to train our multi-class CNN we minimize the cross-entropy loss:
\begin{equation}
L_i = -\log p( y_i|x_i) = -(S(y_i) - \log{\sum_{j}}\exp(S_{j})),
\qquad
L = \frac{1}{N}\sum_{i=1}^{N}L_i
\end{equation}
where $y_i$ is the ground truth label of input $x_i$ and $S(y_i)$ is its corresponding score. The CNN was trained using 140,000 patches for each class.
The networks were trained on a NVIDIA GeForce GTX 980 Ti GPU and implemented using MatConvNet deep learning framework \cite{matconvnet}.
We tried two initialization scenarios for the parameters of the conv. layers: one using random Gaussian distributions ("trained from scratch") and one using transfer learning from the Cifar-10 dataset ("fine-tuned" system). When performing transfer learning, initialization of each channel is pretrained on Cifar-10 dataset separately and the joint fully-connected layers are initialized randomly.
For random initialization (training from scratch) we use learning rate of 0.0001 for the first 30 epochs and decreasing by 1/10 each 10 epochs with total of 50 epochs. Weights are initialized randomly and updated using mini-batches of 128 examples and stochastic gradient descent optimization. Weight decay was chosen to be 0.0001 with momentum of 0.9.
When using transfer learning, convolution layers are initialized with Cifar-10 pre-trained network and learning rate is set to zero. The two last joint fully-connected layers are initialized randomly with learning rate of 1. 

\begin{figure}[t]
  \centering
  \includegraphics[width=0.9\textwidth]{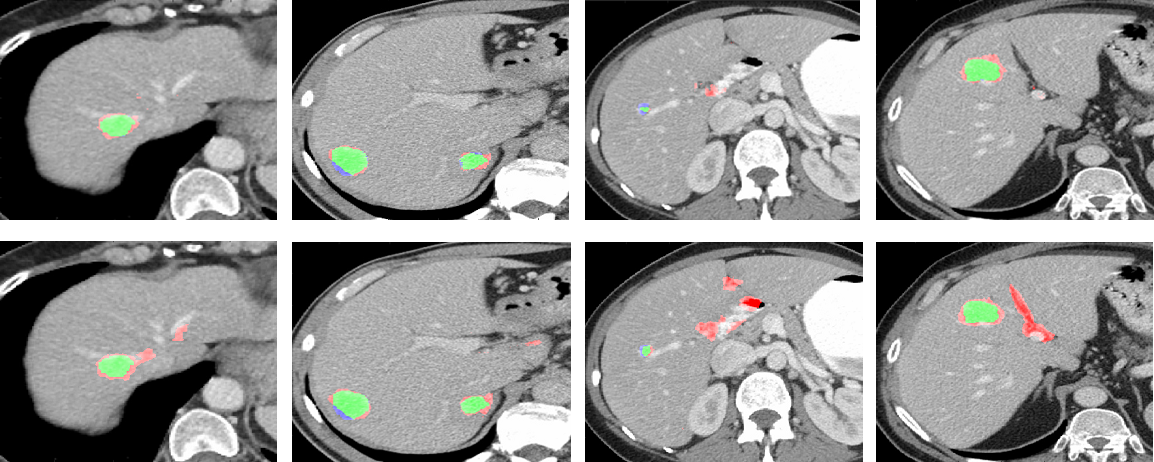}
  \caption{Examples of lesion detection results. First row: Parallel multi-class CNN. Second row: Binary-class CNN. TP marked in green, FP in red and FN in blue.}
\label{Fig:Results_2vs3_classes}
\end{figure}

\section{Experiments and Results}
%
%
%
We evaluated our multi-class CNN system on a liver metastases dataset 
from Sheba Medical Center.
Cases were acquired between 2009 and 2014 using different CT scanners with 0.71-1.17 mm pixel spacing and 1.25-5.0 mm slice thickness. Cases were collected with approval of the institutions Institutional Review Board.
The dataset includes 132 2D liver CT scans with overall of 498 metastases in various shapes, contrast and sizes (5.0-121.0 mm) where each liver contains one or multiple (1-10) lesions. Cases were selected and marked by an expert radiologist.
Images were resized to a fixed pixel spacing of 0.71 mm.

\begin{table*}
\renewcommand{\arraystretch}{1.1}
\caption{Lesion detection performance evaluation: 43 liver dataset; comparison to state-of-the-art \cite{FCN_paper}.}
\begin{center}
\begin{tabular}{ m{5cm}  >{\centering\arraybackslash} m{1.5cm}   >{\centering\arraybackslash} m{1.5cm} } \hline
  Method & TPR  &  FPC \\ \hline\hline
Parallel multi-class CNN  &   98.4\% & 1.0\\
Hierarchical multi-class CNN  &   98.4\% & 0.9\\
Binary-class CNN  &   95.2\% & 1.0\\ \hline
FCN (3 slices) \cite{FCN_paper}  &  88.0\% & 0.74\\
FCN \cite{FCN_paper} &  85.0\% & 1.1\\ \hline
\end{tabular}
\end{center}
\label{Table:Results_43}
\end{table*}

\subsection{Comparison to state-of-the-art}

We first evaluated our proposed approach on the relatively small dataset which was used in \cite{FCN_paper} for comparison purposes.
This dataset includes 20 patients and contains 43 CT liver scans with overall of 68 metastases, where each liver contains 1-3 lesions. Evaluation was performed with 3-fold cross-validation with case separation at the patient level.

Table \ref{Table:Results_43} shows detection performance comparison of our multi-class system to alternative architectures.
We compared our system to the classical detection approach which uses only two classes, lesion and normal tissue, implemented with a {\it {binary-class CNN}}. Results show that our Multi-class approach achieves higher detection performance than using a binary-class CNN. 
Moreover, our proposed system improves over the state-of-the-art FCN system (which was presented in MICCAI 2016 Workshop \cite{FCN_paper}). 
Note that the 3-slice FCN includes also two neighboring slices which we did not use in our implementation.

We note that we obtained comparable results for the hierarchical and parallel multi-class CNNs. The advantage of the parallel scheme is that there is only a single network we need to train.

 Detection results can be seen in Fig. \ref{Fig:Results_2vs3_classes}. The information retained from the sub-categories improves the detection performance and the robustness of the system as compared to the binary-class implementation. It reduces the false-positives in the normal tissue mainly at the liver boundary but also in the interior area of the liver.


\begin{figure}[t]
  \centering
  \includegraphics[width=1\textwidth]{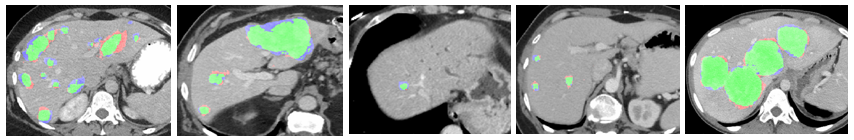}
  \caption{Examples of lesion detection results showing the variability in lesions size, location and shape. TP marked in green, FP in red and FN in blue.}
\label{Fig:results_lesions}
\end{figure}

\begin{table*}[t]
\renewcommand{\arraystretch}{1.1}
\caption{Lesion detection performance evaluation: 132 liver dataset.}
\begin{center}
\begin{tabular}{ m{5cm}  >{\centering\arraybackslash} m{1.5cm}   >{\centering\arraybackslash} m{1.5cm} >{\centering\arraybackslash} m{1.5cm} >{\centering\arraybackslash} m{1.5cm}  } \hline
  Method & \multicolumn{2}{c}{All lesion sizes} & \multicolumn{2}{c}{lesions $>$ 10 mm} \\
  & TPR  &  FPC & TPR & FPC \\ \hline\hline
Parallel multi-class CNN  &   85.9\% & 1.9 &   93.0\% & 1.5\\
Hierarchical multi-class CNN  &   86.8\% & 1.9 & 91.8\% & 1.5\\
Binary-class CNN  &  80.0\% & 2.8 &  70.0\% & 1.9\\ \hline
\end{tabular}
\end{center}
\label{Table:Results_132_all}
\end{table*}

\subsection{Detection performance evaluation}
 We next evaluated the proposed method on a larger dataset that extended the first to include additional variability overall,
 from (institution name withheld).  Evaluation was performed on the 132 livers with 2-fold cross-validation. Table \ref{Table:Results_132_all} shows comparison results of our proposed method to other CNN-based systems. Detection results are shown for the entire dataset, as well as for a subset of the larger lesions $>$ 10 mm, which are the ones mostly recorded by the radiologists and require immediate care.
When applied to all lesion sizes, our proposed method resulted in 85.9\% TPR and 1.9 FP per liver (FPC) while the binary-class CNN resulted in 80\% TPR with 2.8 FPC.
 
Our multi-class hierarchical approach resulted in similar performance of 86.8\% TPR with 1.9 FPC.
Using transfer learning for this task (see section \ref{section:systemArc}) was not as successful as training from scratch (obtaining 82.8\% TPR with 2.54 FPC when applied on all lesion sizes).
The detection performance is higher when excluding small lesions ($< 10 mm$), with 93.0\% TPR and 1.5 FPC using our parallel multi-class system and 91.8\% TPR with 1.56 FPC using our Hierarchical approach. Fig. \ref{Fig:results_lesions} shows detection examples, demonstrating the ability of our system to detect a variety of lesion sizes.
We emphasize that the only difference between the binary-class and the multi-class networks shown in Table \ref{Table:Results_132_all} is the multi-class description of the non-lesion patches at the final softmax layer of the multi-class network. As can be seen from Table \ref{Table:Results_132_all}, this small network architecture difference yields a huge performance difference.

\section{Conclusions}
%
We presented a multi-class patch-based CNN architecture for liver lesion detection in CT images. Our method takes into account the variability of the normal liver tissue and the spatial information retrieved using different scales of view.
We showed that separating the inhomogeneous normal class to sub-categories improves detection performance as compared to a binary-class implementation. Our approach outperformed the state-of-the-art method which is based on a fully convolutional network.
In future work we plan to focus on the lesion class and explore its separation into a dual category: lesion interior and lesion boundary classes, towards improved lesion segmentation. Our approach may generalize to additional medical detection applications. Automated computerized tools, as shown in this work, have the potential in the future to support the radiologists towards improved detection.

%
%

\end{document}